\documentclass[journal,comsoc]{IEEEtran}
\usepackage[T1]{fontenc}
\usepackage{times}
\usepackage{latexsym}

\usepackage{url}
\usepackage{graphicx}  
\usepackage{float}
\usepackage{svg}
\usepackage{enumitem}
\usepackage{multirow}
\usepackage{colortbl}
\usepackage{amssymb}
\usepackage{amsmath}

\ifCLASSINFOpdf
\else
\fi
\usepackage{amsmath}
\usepackage[cmintegrals]{newtxmath}
\begin{document}
%
\title{Multilingual Multi-Domain Adaptation Approaches \\ for Neural Machine Translation}
%
%
%


\author{Chenhui Chu and Raj Dabre
\thanks{C. Chu is with Osaka University, Japan (e-mail: chu@ids.osaka-u.ac.jp).}
\thanks{R. Dabre is with National Institute of Information and Communications Technology, Japan (e-mail: raj.dabre@nict.go.jp).}
}

\maketitle

\begin{abstract}
In this paper, we propose two novel methods for domain adaptation for the attention-only neural machine translation (NMT) model, i.e.,  the Transformer. Our methods focus on training a single translation model for multiple domains by either learning domain specialized hidden state representations or predictor biases for each domain. We combine our methods with a previously proposed black-box method called mixed fine tuning, which is known to be highly effective for domain adaptation. In addition, we incorporate multilingualism into the domain adaptation framework.
Experiments show that multilingual multi-domain adaptation can significantly improve both resource-poor in-domain and resource-rich out-of-domain translations, and the combination of our methods with mixed fine tuning achieves the best performance.
\end{abstract}


%
\IEEEpeerreviewmaketitle

\section{Introduction}
\IEEEPARstart{N}{eural} machine translation (NMT) 
\cite{DBLP:journals/corr/ChoMGBSB14,DBLP:journals/corr/SutskeverVL14,DBLP:journals/corr/BahdanauCB14:original} 
achieves the state-of-the-art translation performance in resource-rich scenarios.
However, domain specific corpora are usually scarce or nonexistent, and thus vanilla NMT performs poorly in such scenarios \cite{DBLP:conf/emnlp/ZophYMK16:original}. 


Domain adaptation techniques leverage out-of-domain data for in-domain translation. In the context of NMT, fine tuning based techniques have been very successful for resource-poor domain translation
\cite{domfast,chu:2017:ACL:original,DBLP:journals/corr/abs-1708-08712:original}. On the other hand, cross-lingual transfer learning methods\footnote{These are also fine tuning techniques. Fine tuning for domain adaptation is a simpler version of cross-lingual transfer learning.} have been successful in improving the performance of low resource languages such as Hausa-English using resource-rich French-English data \cite{DBLP:conf/emnlp/ZophYMK16:original}. 

Most of these methods, however, do not modify the internal structure of the model and rely on black-box approaches. They often incorporating the use of artificial tokens, to improve translation quality. As such, it is not clear, how the artificial tokens affect the learning of the model. Thus, we decide to explicitly model multiple domains by making simple modifications to the decoder. In particular, we either modify the representation of the decoder states before softmax or learn special bias vectors depending on the domain to which the sentence belongs. 

There are studies where either multilingual \cite{DBLP:journals/corr/FiratCB16:original,TACL1081} or multi-domain models \cite{DBLP:journals/corr/abs-1708-08712:original} are trained. However, none that attempt to develop a method and investigate the effect of using both multilingual and multi-domain data, which are more available 
than either
and could be more effective.
In this paper, we present the first work on multilingual and multi-domain NMT models. Our contributions are as follows:
\begin{itemize}
    \item We propose two novel domain adaptation methods that explicitly model domain information for the Transformer: {\it domain specialization} that learns domain specialized hidden state representations, and {\it domain extremization} that learns predictor biases for each domain.
    \item We introduce multilingualism into the {\it fine tuning} \cite{luong2015stanford,sennrich-haddow-birch:2016:P16-11,domspec,domfast}, {\it multi-domain} \cite{domcont}, and {\it mixed fine tuning} \cite{chu:2017:ACL:original} and our proposed methods for domain adaptation. We show that it not only significantly improves translation for an extremely resource-poor domain but also the translations for resource-rich domains. We further combine our methods with mixed fine tuning using multilingual and multi-domain data, and achieves the best results.
    \item We study domain adaptation techniques on the latest attention-only NMT model, the {\it Transformer} \cite{NIPS2017_7181}, which significantly outperforms conventional recurrent neural network (RNN) based NMT models that previous domain adaptation studies work on.
\end{itemize}

\section{Domain Adaptation for NMT}
\label{sec:single_domain}

\subsection{Existing Black-Box Methods}
In this paper, we reproduce previously proposed methods for domain adaptation, which are black-box in nature. As such they are simple and do not need any modifications to the model architecture. In particular, we work with fine tuning, multi-domain and mixed fine tuning,
which uses one out-of-domain corpus to improve the translation of one in-domain corpus. 

\subsubsection{Fine Tuning} We first train an NMT model on a resource-rich out-of-domain corpus (parent model) till convergence, and then resume training on a resource-poor in-domain corpus (child model).

\subsubsection{Multi-Domain} This is motivated by \cite{sennrich-haddow-birch:2016:N16-1,TACL1081}. We simply concatenate the corpora of multiple domains by appending artificial tokens that indicate the domains and by oversampling the corpus of the resource-poor domain following \cite{chu:2017:ACL:original}.

\subsubsection{Mixed Fine Tuning} This was proposed by \cite{chu:2017:ACL:original} and is a combination of the above methods. Instead of fine tuning the out-of-domain model on in-domain data, we fine tune on an in-domain and out-of-domain mixed corpus. This prevents over-fitting and enables smooth domain transition.
Refer to the original paper \cite{chu:2017:ACL:original} for additional details. 


\subsection{Proposed Methods}
Note that for all the proposed methods presented in this section, the resource-poor in-domain corpus is oversampled.
\subsubsection{Domain Specialization (domspec)}
\begin{figure}[t]
    \begin{center}
          \includegraphics[width=0.9\hsize]{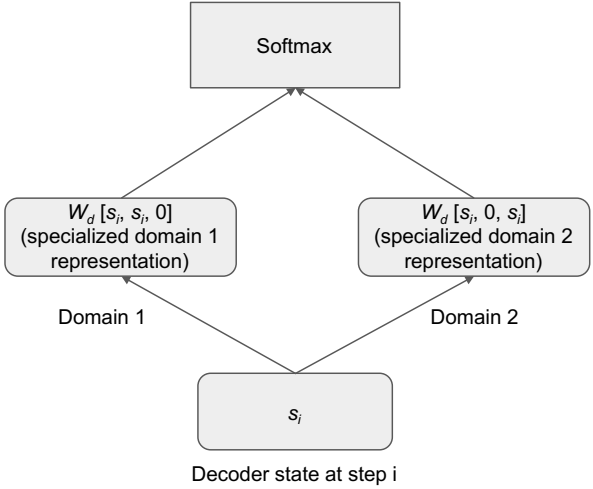}
     \end{center}
    \centering
        \caption{Domain adaptation using a domain specific representation of the decoding states.}
        \label{domspec}
\end{figure}

The motivation of domain specialization is to learn specialized as well as common representations for different domains in a single NMT model. To achieve this, we modify the vanilla NMT model according to the feature replication idea proposed by \cite{daumeiii:2007:ACLMain} for easy domain adaptation.
Figure~\ref{domspec} describes the modification. Accordingly, we perform a simple modification to the decoder state before computing the softmax and call the resultant model as the domain specialization model. Assuming that there are 2 domains, if $s_{i}$ is the state of the decoder for the i'th word to be predicted, the new state passed to the softmax layer is $[s_{i}, s_{i}, 0]$ for words belonging to sentences for the first domain. For the second domain the new state is $[s_{i},  0, s_{i}]$. The $0$ represents a vector which has the same size as $s_{i}$. By doing so, we expect that the decoder will use the first $s_{i}$ to learn some common features for both domains and the remaining $s_{i}$ at the other positions for domain specific features. The resultant decoder state is 3 times the size of the original and thus the softmax layer contains 3 times the number of parameters as the original. In order to reduce the parameter explosion, we down-project $s_{i}$ by a factor of 3 and then perform the replication. For down-projection, we simply perform a linear projection using a weight matrix \(W_d \in \mathbb{R}^{|s_i|\times |s_i|/3}\). This leads to a an insignificant increase in the number of parameters. Note that the input sentences are not pre-pended with an artificial token indicating the domain and hence leave it to feature replication to determine the domain.

\subsubsection{Domain Extremization (domextr)}
\begin{figure}[t]
    \begin{center}
          \includegraphics[width=\hsize]{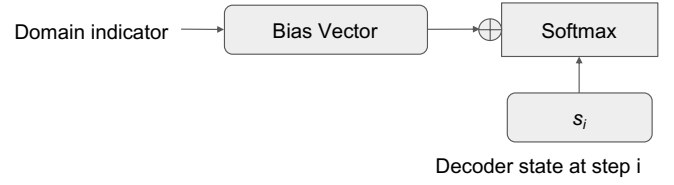}
     \end{center}
    \centering
        \caption{Domain adaptation using a domain specific bias which is added to the logits before computing the probability distribution.}
        \label{domextr}
  \end{figure}

While domain specialization models focus on learning specialized decoder features, domain extremization is motivated by \cite{michel18acl} where special biases are learned for each domain. Figure~\ref{domextr} shows how domain extremization can be performed. The extremization refers to the fact that the softmax decision is guided not by learning special features but by a bias which can help generate extremely different probability distributions. Again, assuming that there are 2 domains, we create 
two extremization vectors $bias_{1}, bias_{2}$. $bias_{1}$ and $bias_{2}$ are vectors of the size of the target sequence vocabulary. The probability distribution to predict the current target word $y_i$ is now computed as:
\begin{equation}
\label{eq:bias}
P(y_{i}|X,y_{<i}) = softmax(W_{t} s_{i} + bias)
\end{equation}
where $X$ is the source sequence, $y_{<i}$ are the previously predicted target words, $s_{i}$ is the current decoder hidden state, \(W_t \in \mathbb{R}^{|s_i|\times |V|}\) is a matrix to map $s_{i}$ to a vector of the size of the vocabulary of the target sequence, and $bias$ denotes either of the two domain bias vectors used for domain extremization.

\subsubsection{Domain Specialization with Extremization (domspecextr)}
This is a simple combination of the domain specialization and extremization methods that incorporates the two domain hidden states $[s_{i}, s_{i}, 0]$ and 
$[s_{i}, 0, s_{i}]$ into Equation \ref{eq:bias} instead of $s_{i}$.
By doing so, we hope that the differentiation of domains will take place before as well as during softmax computation.


\subsubsection{Combination with Mixed Fine Tuning (+MFT)}
Mixed fine tuning is a black-box method and thus is complementary with the above three models. We first train the model with domspec/domextr/domspecextr on the out-of-domain data and then continue training on the combination of the out-of-domain and the in-domain data. Note that artificial domain tags are not used here.

\begin{figure*}[t]
    \begin{center}
          \includegraphics[width=0.75\hsize]{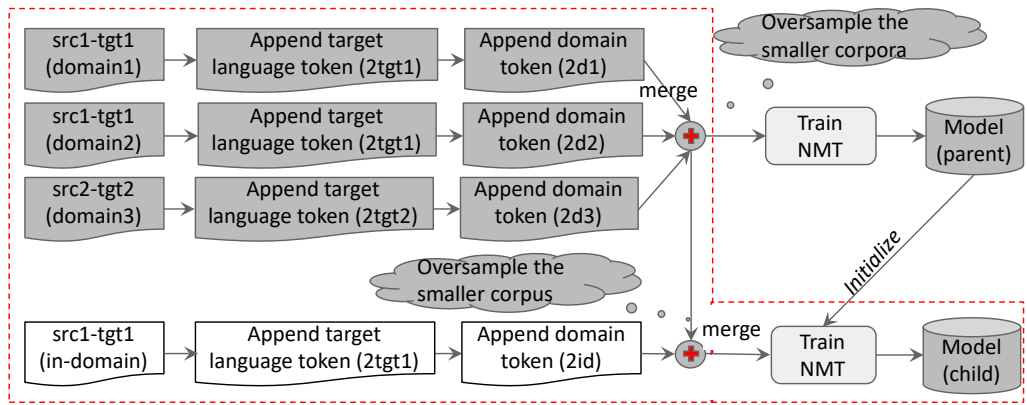}
     \end{center}
    \centering
        \caption{Multilingual and multi-domain adaptation for NMT (the section in the dotted area denotes the multilingual multi-domain method; {2d1 denotes to-domain1 and 2id denotes to-in-domain and so on}). }
        \label{mlnmt}
  \end{figure*}

\section{Multilingual Multi-Domain Adaptation} \label{sec:mlmdnmt}
\label{sec:multi_domain}
We propose to use both multilingual and multi-domain data for domain adaptation.
Figure \ref{mlnmt} gives an overview of our multilingual and multi-domain method. This is a combination of multi-domain \cite{chu:2017:ACL:original} and multilingual NMT \cite{TACL1081}, both of which use artificial tokens to control the target domain or language. Assume that there are multiple source languages, domains and target languages. For simplicity, consider there are two language pairs, src1-tgt1 and src2-tgt2.
For the src1-tgt1 pair, there are one in-domain corpus and two out-of-domain corpora. For the src2-tgt2 pair, there is one out-of-domain corpus.
\subsection{Based on Existing Black-Box Methods}
\subsubsection{Fine Tuning} To train a multilingual out-of-domain parent model (upper part of Figure~\ref{mlnmt}), we append the target language tokens (2tgt1, and 2tgt2)\footnote{Note that when there is only one target language, this language tag can be removed.} and the domain tokens (2d1, 2d2 and 2d3) to the respective corpora; then we merge them by oversampling the smaller corpora and feed this corpus to the NMT training pipeline. 
After that, we fine tune the in-domain model with the parent model.
\subsubsection{Multi-Domain} 
\label{sec:multlingual_multi_domain}
To train a multilingual and multi-domain model, the merged out-of-domain multilingual corpora and the in-domain corpus are further merged into a single corpus by oversampling the smaller corpus. This is then fed to the NMT training pipeline.
\subsubsection{Mixed Fine Tuning} Instead of training a model from scratch, we can apply mixed fine tuning by initializing the multilingual and multi-domain child model using the previous multilingual out-of-domain parent model. This method can reap the benefits of multilingualism as well as mixed fine tuning for domain adaptation.

\subsection{Based on Proposed Methods}
\subsubsection{Domain Specialization (domspec)}
Instead of using two different hidden states to represent different domains, we use multiple different hidden states to represent multiple domains and languages, respective.

\subsubsection{Domain Extremization (domextr)}
Similar to domain specialization, we use multiple domain bias vectors for multiple domains and languages instead of two.

\subsubsection{Domain Specialization with Extremization (domspecextr)}
We use multiple different hidden states and domain bias vectors simultaneously for multiple domains and languages, respective.


\subsubsection{Combination with Mixed Fine Tuning (+MFT)}
The training process is the same as mixed fine tuning using multiple domains and languages, but we use multiple hidden states or domain bias vectors for the decoder.

\section{Experimental Settings}
\label{sec:settings}
\subsection{Multilingual Multi-Domain Settings} \label{sec:mlmdstudy}
We focused on Japanese-English Wikinews translation as the in-domain task. 
This task was conducted on the Japanese-English subset of Asian language treebank (ALT) parallel corpus\footnote{http://www2.nict.go.jp/astrec-att/member/mutiyama/ALT/index.html} 
\cite{KYAWTHU16.435}. This task contains
18088, 1000, and 1018 sentences for training, development, and testing, respectively.
In order to augment the resource-poor ALT-JE in-domain data, we utilized two different out-of-domain corpora with the same source and target languages, and one out-of-domain corpus that only shared the target language with the in-domain corpus. 

The first out-of-domain data was the Kyoto free translation task (KFTT) corpus\footnote{http://www.phontron.com/kftt/} 
\cite{neubig11kftt}. This corpus contains Japanese-English translation that focuses on Wikipedia articles related to the city of Kyoto. This task contains
440288, 1166, and 1160 sentences for training, development, and testing, respectively.
The second out-of-domain data was the spoken domain IWSLT 2017 Japanese-English corpus created by the WIT project \cite{cettoloEtAl:EAMT2012}. This task contains 223108 sentences for training. We used the dev 2010 and test 2010 sets containing 871 and 1549 sentences for development and testing, respectively.
The third out-of-domain data was the spoken domain IWSLT 2015 Chinese-English corpus \cite{cettolo2015iwslt}.
This task contains 209491 sentences for training. 
We used the dev 2010 and test 2010 sets containing 887 and 1570 sentences for development and testing, respectively.

\subsection{MT Systems}
We used the open source implementation of the Transformer model \cite{NIPS2017_7181} in {\it tensor2tensor}\footnote{https://github.com/tensorflow/tensor2tensor} for all our NMT experiments. We used the Transformer because it is the current state-of-the-art NMT model. For training, we used the default model settings corresponding to {\it transformer\_base\_single\_gpu} in the implementation.\footnote{{Note that we used the default adjustment strategy of optimizer's learning rate for all the models for fair comparison and replicability,  leaving the tuning for each model as future work. Also, the number of parameters for all the methods are the same as the {\it transformer\_base\_single\_gpu} in \cite{NIPS2017_7181} because we used the same hyper-parameters and vocabulary sizes.}}
For domain adaptation development, we used the in-domain development data in the fine tuning method, while for all the other methods we used a mix of the in-domain and out-of-domain development data.
We trained the models until convergence.\footnote{When 
there is no change in 0.05 BLEU score over several thousand batches
on the development data.}
For decoding, we averaged the last 20 checkpoints with a beam size of 4 and length penalty $\alpha=0.6$.
We also compared with phrase based SMT (PBSMT) using Moses\footnote{http://www.statmt.org/moses/} \cite{koehn-EtAl:2007:PosterDemo}
for the tasks without domain adaptation as a baseline, because vanilla SMT has been reported to perform better than vanilla NMT in resource-poor translation \cite{DBLP:conf/emnlp/ZophYMK16:original}.
We used default Moses settings\footnote{We trained 5-gram KenLM
language models, used GIZA++
for alignment and MERT \cite{och:2003:ACL} for tuning.} for all our experiments.

For both MT systems, we preprocessed the data as follows: 
Japanese was segmented using JUMAN\footnote{http://nlp.ist.i.kyoto-u.ac.jp/EN/index.php?JUMAN} \cite{kurohashi--EtAl:1994};
English was tokenized and lowercased using the {\it tokenizer.perl} script in Moses;
for Chinese, we used KyotoMorph\footnote{https://bitbucket.org/msmoshen/kyotomorph-beta} for segmentation. 
In order to reduce the number of out of vocabulary words in NMT, we pre-processed the corpora using the default sub-word segmentation mechanism, which is a part of tensor2tensor. For all our NMT experiments, we set the source and target vocabularies sizes to 32000 sub-words. We followed the vocabulary acquisition methods of \cite{chu:2017:ACL:original}
for all the domain adaptation methods,
but used a vocabulary random mapping from Chinese to Japanese following \cite{DBLP:conf/emnlp/ZophYMK16:original} when fine tuning on the ALT-JE data using the IWSLT-CE data. 
\section{Results}

\begin{table}[t]
 \small
 \begin{center}
 \begin{tabular}{@{}l|r|r|r|r@{}}\hline 
System & ALT-JE & KFTT-JE & IWSLT-JE & IWSLT-CE \\ \hline 
 ALT-JE SMT & {\bf 11.03} & {(2.16)} & {( 1.93)} & {(0.20)} \\ 
 ALT-JE NMT & 8.47 & {(1.66)} & {(1.75)} & {(0.28)} \\ 
 KFTT-JE SMT & {(5.59)} & 19.16 & {(2.93)} & {(0.65)} \\ 
 KFTT-JE NMT & {(5.44)} & \bf{27.19} & {(2.15)} & {(0.00)} \\ 
 IWSLT-JE SMT & {(6.32)} & {(1.98)} & 7.98 & {(1.02)} \\ 
 IWSLT-JE NMT & {(10.65)} & {(2.30)} & \bf{11.09} & {(0.62)} \\ 
 IWSLT-CE SMT & {(0.55)} & {(0.73)} & {(0.83)} & 12.73 \\ 
 IWSLT-CE NMT & {(0.21)} & {(0.32)} & {(0.46)} & \bf 16.89 \\ \hline
 \end{tabular}%
 \end{center}
  \caption{\label{table:wo-domain} Translation results (BLEU-4 scores) without domain adaptation. The numbers in {\bf bold} indicate the best scores using the same data. Scores in parentheses are listed for reference, and they show low performance due to domain/language mismatch.
 }
 \end{table}
 

\begin{table}[t]
\small
\begin{center}
\begin{tabular}{l|l|l}\hline 
\bf No. & \bf KFTT-JE+ALT-JE & \bf ALT-JE  \\ \hline 
1 & concat & 21.62  \\
2 & fine tuning & 16.50  \\
3 & multi-domain & 16.24 \\
4 & mixed fine tuning & 21.74 \\
5 & domspec & 16.90 \\
6 & domspec+MFT & 22.43 \\
7 & domextr & 16.96 \\
8 & domextr+MFT & \bf 23.05\dag \\
9 & domspecextr & 16.29 \\
10 & domspecextr+MFT & 22.07 \\\hline
\bf No. & \bf IWSLT-JE+ALT-JE & \bf ALT-JE  \\ \hline 
11 & concat & 19.11 \\
12 & fine tuning & {\bf 20.42} \\
13 & multi-domain & 14.92 \\
14 & mixed fine tuning & 19.76 \\
15 & domspec & 13.88 \\
16 & domspec+MFT & 18.70 \\
17 & domextr & 14.28 \\
18 & domextr+MFT & 18.86 \\
19 & domspecextr & 13.16 \\
20 & domspecextr+MFT & 18.83 \\\hline
\bf No. & \bf IWSLT-CE+ALT-JE & \bf ALT-JE  \\ \hline 
21 & concat & 18.01 \\
22 & fine tuning & 16.32 \\
23 & multi-domain & 16.08 \\
24 & mixed fine tuning & {\bf 19.10} \\ 
25 & domspec & 12.18  \\
26 & domspec+MFT & 17.10 \\
27 & domextr & 11.52 \\
28 & domextr+MFT & 16.47 \\
29 & domspecextr & 11.21 \\
30 & domspecextr+MFT & 16.77 \\\hline
\end{tabular}%
\end{center}
\caption{\label{table:one-domain-in} Domain adaptation results (BLEU-4 scores) with one out-of-domain corpus for ALT-JE using KFTT-JE, IWSLT-JE and IWSLT-CE. The numbers in {\bf bold} indicate the best scores using the same data. The numbers marked with ``\dag'' indicate that the results using our proposed methods are significantly better (p \textless 0.05) than the existing black-box methods for the same data setting. 
}
\end{table}

\subsection{Without Domain Adaptation} Table \ref{table:wo-domain} shows the vanilla PBSMT and NMT results. 
Each system was trained for a particular MT task without any domain adaptation. 
We can see that SMT performs better for the in-domain translation and using out-of-domain models for in-domain translation shows poor performance. It is clear that a domain or language mismatch leads to poor translation quality (i.e., scores in parentheses in Table \ref{table:wo-domain}) and thus do not report the BLEU scores when there is a mismatch in the domain adaptation experiments (i.e., Sections \ref{sec:one_ood_res}, \ref{sec:multi_res}, and \ref{sec:feasibility_res}).

\subsection{Adaptation with One Out-of-Domain Corpus}
\label{sec:one_ood_res}
Table \ref{table:one-domain-in} shows the in-domain results for domain adaptation using only one out-of-domain corpus.\footnote{Refer to Table \ref{table:one-domain-out} the appendix for the out-of-domain results.}
We also conducted NMT experiments that simply concatenated the corpora, denoting as ``concat'' in the table.
We can see that using a single out-of-domain corpus improves the in-domain translation.
Although the corpus size of KFTT-JE is two times larger than that of IWSLT-JE, the performance is not better besides using mixed fine tuning. This indicates that the size of the out-of-domain corpus is not the only decisive factor for domain adaptation but the method also matters, which also can be indicated in the multiple domain and multilingual multi-domain adaptation results. 

Unfortunately, domspec, domextr and domainspecextr are not always significantly better than previous methods. We suspect the reason for this is due to the small amount of in-domain data, making it difficult to learn either the domain specific hidden states or biases. 
The proposed methods learn models from scratch and the small in-domain data makes it difficult to learn the in-domain models, while both fine tuning and mixed fine tuning depend on transfer learning.\footnote{Given a sufficiently large in-domain corpus, it is possible for our methods to beat fine tuning and mixed fine tuning. But this could not provide solutions for small in-domain translation.}
However, combining them with mixed fine tuning significantly improves BLEU scores, which also achieves the best in-domain performance using KFTT-JE data. We believe the reason for this is the robustness of the mixed fine tuning model that is pre-trained on the out-of-domain only, 
making our methods learn better domain representations. However, different from KFTT-JE, for the IWSLT-JE and IWSLT-CE data, combining the proposed methods with mixed fine tuning does not show better performance than the existing black-box methods. We believe that the main reason is the size difference of the out-of-domain corpora. 

An interesting observation is that cross-lingual transfer across domains shows comparable results compared to using the out-of-domain corpus from the same language pair when the two corpora have similar characteristics, i.e., IWSLT-CE v.s. IWSLT-JE. This means that, in cases where out-of-domain corpora for the same language pair are not available, using out-of-domain corpora that share only the target language is also useful.

\begin{table}[t]
\small
\begin{center}
\begin{tabular}{l|l|r}\hline 
\multicolumn{3}{l}{\bf Multiple Out-of-domain Corpora} \\ \hline 
\bf No. & \bf KFTT-JE+IWSLT-JE+ALT-JE & \bf ALT-JE \\ \hline 
1 & concat & 23.41 \\
2 & fine tuning & 23.09 \\
3 & multi-domain & 20.77 \\
4 & mixed fine tuning & 24.29 \\
5 & domspec & 19.64 \\
6 & domspec+MFT & 23.62 \\
7 & domextr & 21.04 \\
8 & domextr+MFT & \bf 24.41 \\
9 & domspecextr & 18.82 \\
10 & domspecextr+MFT & 23.93 \\ \hline 
\multicolumn{3}{l}{\bf Multilingual Single Out-of-domain Corpora} \\ \hline 
\bf No. & \bf IWSLT-JE+IWSLT-JE+ALT-JE & \bf ALT-JE \\ \hline 
11 & concat & 19.91 \\
12 & fine tuning & {21.22} \\
13 & multi-domain & {17.99} \\
14 & mixed fine tuning & {19.35} \\ 
15 & domspec & 16.01 \\
16 & domspec+MFT & 20.53 \\
17 & domextr & 16.84 \\
18 & domextr+MFT & \bf 21.42 \\
19 & domspecextr & 16.09 \\
20 & domspecextr+MFT & 20.67 \\\hline 
\multicolumn{3}{l}{\bf Multilingual Multi-Domain Adaptation} \\ \hline 
\bf No. & \bf KFTT-JE+IWSLT-JE+IWSLT-JE+ALT-JE & \bf ALT-JE \\ \hline 
21 & concat & 23.71 \\
22 & fine tuning & 23.42 \\
23 & multi-domain & 21.97 \\
24 & mixed fine tuning & 24.04 \\
25 & domspec & 20.51 \\
26 & domspec+MFT & 23.25 \\
27 & domextr & 21.62 \\
28 & domextr+MFT & {\bf 24.76}\ddag \\
29 & domspecextr & 20.90 \\
30 & domspecextr+MFT & 23.46 \\\hline 
\end{tabular}%
\end{center}
\caption{\label{table:multi-in} Multilingual multi-domain adaptation results (BLEU-4 scores) for ALT-JE using KFTT-JE, IWSLT-JE and IWSLT-CE. The numbers in {\bf bold} indicate the best scores using the same data. The numbers marked with ``\ddag'' indicate the best score over all data settings. 
}
\end{table}

\subsection{Multilingual Multi-Domain Adaptation}
\label{sec:multi_res}
Table~\ref{table:multi-in} shows the in-domain results for multilingual multi-domain adaptation,\footnote{Refer to Table \ref{table:multi-out} the appendix for the out-of-domain results.} where
the above sub-table contains results for using two out-of-domain corpora in the same language pair;
the middle sub-table contains results for using multilingual single domain corpora;
the below sub-table is for multilingual multi-domain adaptation where multiple out-of-domain corpora with different source languages are used. Again, ``concat'' denotes the NMT baselines that simply concatenate the corpora.


We can see that increasing the number of domains further boosts the in-domain performance.
Although data from mixing different domains increases the difficulty of training a single NMT model,\footnote{This is due to the increase of domain diversity such as vocabularies and styles. This involves learning domain specialized representations, which increases the difficulty.} 
the increase of data size leads to better parent models that consequently improves the in-domain translation.
Combination of our proposed methods with mixed fine tuning again performs the best.

Domain adaptation with multilingual single out-of-domain corpora also perform better than using one out-of-domain corpus with large improvements.
Although the source language is different, the decoder model is boosted by mixing IWSLT-JE and IWSLT-CE, which we think is the main reason for improvement. Similarly, the combination of our proposed methods with mixed fine tuning outperforms the other methods.

For multilingual multi-domain adaptation, we can see that
using multilingualism together with multi-domain data shows the best results.
From Table~\ref{table:multi-in}, we can see that domextr+MFT always favors ALT-JE, for which the least amount of data is available. As such, future methods for improving the translation quality for the language pairs with the smallest datasets should incorporate domain extremization methods. 

\begin{table*}[t]
\small
\begin{center}
\begin{tabular}{@{}l|l|r|r|r|r@{}}\hline 
 No. & System   & ALT-EJ (in-domain) & KFTT-EJ & IWSLT-EJ & IWSLT-EC \\ \hline
1 & Vanilla SMT  & 15.1  & 23.8 & 8.0 & 9.5 \\ 
2 & Vanilla NMT  & 13.2  & \bf 32.7 & 9.5 & 11.5 \\ \hline
3 & IWSLT-EC (multi-domain, cross-lingual)  & 14.9 & - & - & 11.2 \\ 
4 & IWSLT-EC (mixed fine tuning, cross-lingual)  & 19.5 & - & - & 11.1 \\ 
5 & IWSLT-EJ\_IWSLT-EC (multi-domain) & 20.0 & - & 11.4 & 12.8 \\ 
6 & IWSLT-EJ\_IWSLT-EC (mixed fine tuning)  & 24.7 & - & 11.4 & 12.7 \\ 
7 & KFTT-EJ\_IWSLT-EJ\_IWSLT-EC (multi-domain) & 23.5 & 32.1 & \bf 12.2 & 13.1 \\ 
8 & KFTT-EJ\_IWSLT-EJ\_IWSLT-EC (mixed fine tuning) & {\bf 27.4} & 31.1 & 12.0 & \bf 13.3 \\ \hline
\end{tabular}%
\end{center}
\caption{Multilingual multi-domain adaptation results (BLEU-4 scores) for ALT-EJ using IWSLT-EC, IWSLT-EJ and KFTT-EJ where multiple target languages exist. The numbers in {\bf bold} indicate the best scores over all settings.}
\label{multitargetmft}
\end{table*}

It turns out that when we use multiple out-of-domain corpora, domextr works consistently better than domspec for in-domain translation. We think the reason for this is that when the number of out-of-domain corpora increases, the in-domain representation ability of domspec decreases due to down-projection, but for domextr it remains the same. Unfortunately, domspecextr that combines domain specialization and extremization approaches fails to improve beyond the individual methods. Domspecextr combines both, which significantly increases the number of new parameters, making the model harder to learn.

We randomly investigated 50 ALT-JE  translations by domextr+MFT. We found that the translation quality reaches to a practical level after our adaptation. The use of multi-domain data improves the translation of not only common words (23/50 sentences) but also domain specific terminologies (7/50 sentences). Multilingualism is mainly helpful for common word translation (11/50 sentences), but it also leads to some noise (4/50 sentences) for terminology translation due to the vocabulary mismatch of low frequent terms. Using both multilingual and multi-domain data improves common word translation more than using only multi-domain data 
(31/50 sentences). 

\subsection{Feasibility in Multiple Target Language Scenarios}
\label{sec:feasibility_res}
We also conduced experiments when there are multiple target languages.
We flipped the translation directions and trained multilingual, multi-domain models for ALT-EJ, KFTT-EJ, IWSLT-EJ and IWSLT-EC. As the fine tuning, domain specialization, and extremization methods were not designed for multiple target languages, we only experimented with the basic multilingual multi-domain models (see Section \ref{sec:multlingual_multi_domain}) in combination with mixed fine tuning. The datasets and model training settings were the same as mentioned in the experimental setting section. 
Table~\ref{multitargetmft} shows the results. Lines 1 and 2 give the scores of the vanilla SMT and NMT models for the tasks.
Lines 3 to 8 contain the results of 3  data settings for adaptation (IWSLT-EC for ALT-EJ, IWSLT-EJ\_IWSLT-EC for ALT-EJ, and KFTT-EJ\_IWSLT-EJ\_IWSLT-EC for ALT-EJ) where 2 target languages are available. We skipped other data settings, because they do not involve multiple target languages. 

We can see that the results are similar to the ones with one target language that: IWSLT-EC also helps improve ALT-EJ although it is cross-lingual. Multilingual and multi-domain adaptation further significantly improves the performance of the in-domain ALT-EJ translation. Using more data performs better (i.e., KFTT-EJ\_IWSLT-EJ\_IWSLT-EC  v.s. IWSLT-EJ\_IWSLT-EC). Multilingual and multi-domain adaptation also improves the translation quality of 
out-of-domain IWSLT-EJ and IWSLT-EC, but not KFTT-EJ. We believe the reason for this is because KFTT-EJ already has enough training data. Mixed fine tuning performs significantly better than multi-domain, which is a new finding in multiple target language translation settings.

\section{Related Work}
Kim et al. \cite{kim-etal-2016-frustratingly} extended the feature replication idea of \cite{daumeiii:2007:ACLMain} for neural domain adaptation on slot tagging tasks, where they use one RNN layer for common representations and additional multiple RNN layers for domain specific representations. In contrast, our domspec method implements the feature replication idea in the decoding state of NMT. Michel and Neubig \cite{michel18acl} conducted adaptation for each speaker in the TED tasks by leaning a speaker bias vector, while our domextr method learns a bias for each domain.
Thompson et al. \cite{thompson-etal-2018-freezing} analyzed the effect of each component in fine tuning based NMT adaptation. Britz et al. \cite{britz-le-pryzant:2017:WMT} proposed to use a feed-forward network as a domain discriminator for NMT domain adaptation, which is jointly optimized with NMT. 

Fine tuning has also been explored for domain adaptation for other NLP tasks using neural networks (NN). Mou et al. (\cite{mou-EtAl:2016:EMNLP2016}) used fine tuning for both equivalent/similar tasks but with different data sets and different tasks but that share the same NN architecture. They found that the effectiveness of fine tuning depends on the relatedness of the tasks. Tag based NMT has also been shown to be effective for other sub tasks of NMT. Sennrich et al. (\cite{sennrich-haddow-birch:2016:N16-1}) tried to control the politeness of translations by appending a politeness tag to the source side language that uses honorific. Johnson et al. (\cite{gnmt16multi}) mixed different language pairs by appending a target language tag to the source text of each language for training a multilingual NMT system. 

Monolingual corpora are widely used for SMT. In SMT, they are used for training a LM, and the LM is used as a feature for the decoder in a log-linear model \cite{koehn-EtAl:2007:PosterDemo}{, \cite{och-ney:2002:ACL}}. {In-domain monolingual data has been used for NMT in other ways \cite{C18-1111}. Currey et al. (\cite{currey-micelibarone-heafield:2017:WMT}) copied the target monolingual data to the source side and used the copied data for training NMT. Domhan and Hieber (\cite{domhan-hieber:2017:EMNLP2017}) proposed using target monolingual data for the decoder with LM and NMT multitask learning. Zhang and Zong (\cite{zhang-zong:2016:EMNLP2016}) used source side monolingual data to strengthen the NMT encoder. Cheng et al. (\cite{cheng-EtAl:2016:P16-1}) used both source and target monolingual data for NMT trough reconstructing the monolingual data with an autoencoder. We leave the comparison with these recently proposed methods as a topic for future work.}

\section{Conclusion}
In this paper, we proposed two novel domain adaptation methods that explicitly model domain information in the decoder. Combining with mixed fine tuning, our methods achieved the best translation performance. Furthermore, we proposed to use both multilingual and multi-domain data for improving in-domain NMT. We also explored the feasibility of mixed fine tuning in a multiple target languages scenario.
Experiments on the Transformer showed the effectiveness of multilingual and multi-domain adaptation. As future work, we plan to experiment on more domains and language pairs, and much larger datasets to compare with state-of-the-art results.

\appendices
\label{app}
\section{Out-of-domain Translation Results}
Table \ref{table:one-domain-out} shows the out-of-domain results after domain adaptation with only one out-of-domain corpus. We can see that after combining with mixed fine tuning, our proposed methods also improve out-of-domain translation with the exception of IWSLT-CE. The reason for IWSLT-CE not improving as much as the JE language pairs is because this translation direction does not benefit from additional source side Chinese corpus.

\begin{table}[t]
\small
\begin{center}
\begin{tabular}{l|l|l}\hline 
\bf No. & \bf KFTT-JE+ALT-JE &\bf KFTT \\ \hline 
1 & concat & 26.20 \\
2 & fine tuning & {4.25}\\
3 & multi-domain & 23.45\\
4 & mixed fine tuning & 25.08\\
5 & domspec & 27.07\\
6 & domspec+MFT & 27.48\\
7 & domextr & 26.11\\
8 & domextr+MFT & \bf 27.98\dag\ddag \\
9 & domspecextr & 26.79 \\
10 & domspecextr+MFT & 27.44\\\hline
\bf No. & \bf IWSLT-JE+ALT &\bf IWSLT-JE \\ \hline 
11 & concat & 10.65 \\
12 & fine tuning & {6.13} \\
13 & multi-domain & 10.33\\
14 & mixed fine tuning & 11.23 \\
15 & domspec & 10.99  \\
16 & domspec+MFT & \bf 11.36  \\
17 & domextr & 10.78 \\
18 & domextr+MFT & 11.23 \\
19 & domspecextr & 10.97 \\
20 & domspecextr+MFT & 11.33  \\\hline
\bf No. & \bf IWSLT-CE+ALT &\bf IWSLT-CE \\ \hline 
21 & concat & 16.37 \\
22 & fine tuning & 0.00 \\
23 & multi-domain & \bf 16.61 \\
24 & mixed fine tuning & 16.28 \\ 
25 & domspec & 15.91 \\
26 & domspec+MFT & 16.19 \\
27 & domextr & 15.40 \\
28 & domextr+MFT & 15.59 \\
29 & domspecextr & 15.30 \\
30 & domspecextr+MFT & 15.40  \\\hline
\end{tabular}
\end{center}
\caption{\label{table:one-domain-out} Out-of-domain results (BLEU-4 scores) after domain adaptation with one out-of-domain corpus for ALT-JE using KFTT-JE, IWSLT-JE and IWSLT-CE. The numbers in {\bf bold} indicate the best scores using the same data. The numbers marked with ``\dag'' indicate that the results using our proposed methods are significantly better (p \textless 0.05) than the existing black-box methods for the same data setting. The numbers marked with ``\ddag'' indicate the best score over all data settings. Note that fine tuning was conducted on the in-domain data only and thus using this in-domain model to translate out-of-domain/language data shows significantly low performance. 
}
\end{table}

Table \ref{table:multi-out} shows the out-of-domain results after multilingual multi-domain domain adaptation. We can see that the combination of our proposed methods with mixed fine tuning performs the best. For multilingual multi-domain adaptation, using multilingualism together with multi-domain data shows the best results for two out-of-domain translations, i.e., IWSLT-JE and IWSLT-CE. We also observe that combining our proposed methods with MFT has a positive impact on the relatively resource-rich IWSLT-JE and IWSLT-CE translation directions. In contrast, vanilla MFT does not achieve this kind of improvement. According to us, our methods learn better specialized representations or biases when provided with additional types of domains. 
Our results indicate that it is possible to package multiple language pairs and domains into a single NMT model with significant improvement for both in-domain and out-of-domain translations.

\begin{table*}[t]
\small
\begin{center}
\begin{tabular}{l|l|r|r|r}\hline 
\multicolumn{5}{l}{\bf Multiple Out-of-domain Corpora} \\ \hline 
\bf No. & \bf KFTT-JE+IWSLT-JE+ALT-JE &\bf KFTT-JE &\bf IWSLT-JE &\bf IWSLT-CE \\ \hline 
1 & concat & 25.99 & 12.90 & - \\
2 & fine tuning & {6.47} & {7.40} & - \\
3 & multi-domain & 24.33 & 11.98 & - \\
4 & mixed fine tuning & 25.33 & 12.33 & - \\
5 & domspec & 26.82 & 13.63 & - \\
6 & domspec+MFT & 27.11 & \bf 13.68\dag & - \\
7 & domextr & 26.34 & 13.34 & - \\
8 & domextr+MFT & \bf 27.33\dag & 13.57 & - \\
9 & domspecextr & 26.69 & 13.19 & - \\
10 & domspecextr+MFT & 26.67 & 13.66 & - \\ \hline 
\multicolumn{5}{l}{\bf Multilingual Single Out-of-domain Corpora} \\ \hline 
\bf No. & \bf IWSLT-JE+IWSLT-JE+ALT-JE &\bf KFTT-JE &\bf IWSLT-JE &\bf IWSLT-CE \\ \hline 
11 & concat & - & 11.23 & 15.06 \\
12 & fine tuning & - & {6.78} & {7.51} \\
13 & multi-domain & - & {11.01} & {15.39} \\
14 & mixed fine tuning & - & {10.91} &  {14.89} \\ 
15 & domspec & - & 12.72 & \bf 17.42\dag \\
16 & domspec+MFT & - & \bf 13.03\dag & 17.39 \\
17 & domextr & - & 12.23 & 16.67 \\
18 & domextr+MFT & - & 12.31 & 16.76 \\
19 & domspecextr & - & 12.57 & 16.91 \\
20 & domspecextr+MFT & - & 12.87 & 17.49 \\\hline 
\multicolumn{5}{l}{\bf Multilingual Multi-Domain Adaptation} \\ \hline 
bf No. & \bf KFTT-JE+IWSLT-JE+IWSLT-JE+ALT-JE &\bf KFTT-JE &\bf IWSLT-JE &\bf IWSLT-CE \\ \hline 
21 & concat & 26.58 & 13.03 & 17.84 \\
22 & fine tuning & {7.20} & {7.66} & {9.12} \\
23 & multi-domain & 26.10 & 12.52 & 16.42 \\
24 & mixed fine tuning & 26.00 & 11.77 & 16.40 \\
25 & domspec & 26.37 & 13.77 & 18.38 \\
26 & domspec+MFT & 26.34 & \bf 13.96\dag\ddag & 18.37 \\
27 & domextr & 26.33 & 13.09 & 18.09 \\
28 & domextr+MFT &  26.50 & 13.60 & \bf 18.44\dag\ddag \\
29 & domspecextr & 26.16 & 13.35 & 17.73 \\
30 & domspecextr+MFT & \bf 26.83 & 13.93 & 18.23 \\\hline 
\end{tabular}%
\end{center}
\caption{\label{table:multi-out} Out-of-domain results (BLEU-4 scores) after multilingual multi-domain adaptation results (BLEU-4 scores) for ALT-JE using KFTT-JE, IWSLT-JE and IWSLT-CE. The numbers in {\bf bold} indicate the best scores using the same data. The numbers marked with ``\dag '' indicate that the results using our proposed methods is significantly better (p \textless 0.05) than the existing black-box methods for the same data setting. The numbers marked with ``\ddag'' indicate the best score over all data settings. Note that fine tuning was conducted on the in-domain data only and thus using this in-domain model to translate out-of-domain/language data shows low performance. 
}
\end{table*}


\section*{Acknowledgment}

This work was supported by Grant-in-Aid for Research Activity Start-up \#17H06822, JSPS.

\ifCLASSOPTIONcaptionsoff
  \newpage
\fi

\bibliography{IEEEtran}
\bibliographystyle{IEEEtran}




\end{document}